\pdfoutput=1

\documentclass[11pt]{article}

\usepackage[final]{acl}

\usepackage{times}
\usepackage{latexsym}
\usepackage{amsmath}

\usepackage[T1]{fontenc}

\usepackage[utf8]{inputenc}

\usepackage{microtype}

\usepackage{inconsolata}

\usepackage{graphicx}

\title{Occam's Razor and Bender and Koller's Octopus}

\author{Michael Guerzhoy \\
  University of Toronto \\
  \texttt{guerzhoy@cs.toronto.edu}} 
  
\begin{document}
\maketitle
\begin{abstract}
We discuss the teaching of the discussion surrounding Bender and Koller's prominent ACL 2020 paper, ``Climbing toward NLU: on meaning form, and understanding in the age of data"~\cite{bender2020climbing}.

We present what we understand to be the main contentions of the paper, and then recommend that the students engage with the natural counter-arguments to the claims in the paper.

We attach teaching materials that we use to facilitate teaching this topic to undergraduate students.
\end{abstract}

\section{Introduction}

The claim in Bender and Koller (B\& K)'s argument in~\cite{bender2020climbing} is that a being that only has access to the \text{form} of the communication -- e.g., an intelligent octopus that taps into only the submarine signals that encode accounts of the events above the sea that two people on land send each other -- will not be able to ``understand" what is happening above sea-level, lacking the semantics of the Morse code that was used to communicate the events transpiring above the seas. Koller and Bender argue that even if the octopus can send messages based on the patterns it sees that would be understood by the humans and the humans would be fooled into thinking they are reading messages from another human, the shallow understanding of the octopus would necessarily be revealed when trying to pretend to answer more complicated queries.

Implicit in the argument is that the intelligent octopus is analogous to a Large Language Model (LLM), akin to GPT-2~\cite{brown2020language}, GPT-4~\footnote{https://openai.com/index/gpt-4-research/}, Claude~3~\footnote{https://www.anthropic.com/news/claude-3-family}, or Meta Llama 3~\footnote{https://llama.meta.com/llama3/}, and that such LLMs would not be able to truly understand natural language in the same way that B\& K's octupus will not.

In this paper, we present a lecture + activity that challenges B\& K's argument. Students will engage with B\& K's argument and with the counterargument, and come away with their own conclusions

\section{Building theory from data}
The scientific process itself can be analogized to a B\&K octopus observing data they don't understand.

For example, astronomers observe and try to predict the motions of heavenly bodies, initially with no mechanistic understanding of why the stars appear to move the way they do. Historically, astronomers came up with multiple incorrect theories for why the heavenly bodies move the way they do (notably, the family of geocentric models). Astronomers used ``epicycles" as a way to align predictions with their model, at the expense of parsimony~\cite{duhem2015save}.

Historically, Copernicus' models used epicycles~\cite{riccioli2023almagestum}. The simplest possible Copernican model with no epicycles would be much simpler but would predict worse than the state-of-the-art model at the time.

Note that, unlike the astronomers, the octopus cannot \textit{interact} with the world -- he cannot influence what observations are made (at least before he starts communicating with the astronomers). This can influence how fast the octopus can ``converge." Historically, much of the data used by Kepler was previously collected by Tycho Brahe.

\subsection{Occam's razor}
Occam's razor -- the principle that, all things being equal, we should prefer the simpler theory -- can help select the better scientific theory. For example, the B\& K octopus might consider all possible theories of the world over the sea, and settle on the simplest one that explains the communications the octopus decodes.

\section{Can the B\& K octopus learn science}
Every student can make their own conclusions, but ours is that it's not in principle impossible for that to happen (or if it is impossible, we don't have a clear reason to think so). The success of LLMs on tasks that require some level of world-theory-building such as the addition of integers task~\cite{lee2023teaching}, predicted to be impossible by \citet{bender2020climbing} (see Appendix B), indicates that if there are barriers to learning world models from observational data, they are not well-understood. Our view is that the prediction by B\&K that a pure LLM could not learn to do arithmetic is due to insufficiently accounting for the possibility of using inductive biases to build a model of the data that corresponds to the world that the data is describing.

\section{Materials}
We provide slides we used in class to follow up the class's reading of \citet{bender2020climbing}. We also provide the following guiding questions

\begin{enumerate}
    \item If the octopus observes different content in messages when it's dark vs. when it's light, what can the octopus possibly conclude about language?
    \item Describe how the octopus might use tides to infer words that have to do with tides
    \item Describe how the octopus might decode conversations about physics based on the conversations about tides -- perhaps building up from observations of tides, stars, etc.
    \item If you assume no ``cheating" such as jointly observing tides, might you imagine conversations that involve physical and mathematical constant like $G$ and $\pi$ playing a similar role?
    \item Explain why without Occam's razor, the Octopus will have a practically infinite number of theories about what the two humans could be talking about
    \item What might be some insurmountable challenges for the octopus in the quest to understand the meaning of the cable signals? How 
    \item Consider the claim from the original paper that arithmetic is not learnable by form alone: where might that argument have gone wrong?
\end{enumerate}

\section{Additional materials}
Julian Michael, \textit{To Dissect An Octopus} \url{https://julianmichael.org/blog/2020/07/23/to-dissect-an-octopus.html} provides an excellent overview.

\section{Conclusion}
Many students in NLP would be familiar with B\& K's argument, but have probably not engaged in the critical analysis of the arguments. We provide materials for critically analyzing the arguments made by B\& K. We focus on the counterarguments since the argument itself is ably presented by the original authors. We provide slides introducing the B\&K argument to the best our ability as well.

Many (though not all) students are captivated by the debate. We find that the structure provided by the guiding questions helps in our lectures.

\section{Teaching materials}
\textbf{Slides}: \url{https://github.com/guerzh/octopus}

\noindent \textbf{Video lecture}: \url{https://youtu.be/6QVjGF_J7I0}

\bibliography{acl_latex}

\end{document}